\documentclass[final]{cvpr}

\usepackage{times}
\usepackage{epsfig}
\usepackage{graphicx}
\usepackage{amsmath}
\usepackage{amssymb}
\usepackage{bm}
\usepackage{multirow}

\usepackage[pagebackref=true,breaklinks=true,colorlinks,bookmarks=false]{hyperref}

\def\cvprPaperID{3494} 
\def\confYear{CVPR 2021}

\begin{document}

\title{Learning Compositional Radiance Fields of Dynamic Human Heads}

\author{
Ziyan Wang$^{1,3}$~~~~~ 
Timur Bagautdinov$^{3}$~~~~~
Stephen Lombardi$^{3}$~~~~~
Tomas Simon$^{3}$\\
Jason Saragih$^{3}$~~~~~
Jessica Hodgins$^{1,2}$~~~~~
Michael Zollhöfer$^{3}$

\vspace{0.1cm} \\ 
$^{1}$Carnegie Mellon University~~~
$^{2}$Facebook AI Research~~~
$^{3}$Facebook Reality Labs

\vspace{0.1cm} \\
\href{https://ziyanw1.github.io/hybrid_nerf/}{https://ziyanw1.github.io/hybrid\_nerf/}
}


 \newcommand{\TB}[1]{{\color{green}{\bf TB: #1}}}
  
 \newcommand{\tb}[1]{{\color{green} #1}}
 
 \newcommand{\TS}[1]{{\color{BurntOrange}{\bf TS: #1}}}
 
 \newcommand{\ZW}[1]{{\color{blue}{\bf ZW: #1}}}
 
 \newcommand{\SL}[1]{{\color{gray}{\bf SL: #1}}}
 
 \definecolor{mzcolor}{RGB}{255,50,00}
 \newcommand\MZ[1] {\textbf{\textcolor{mzcolor}{MZ: #1}}}
 
 \newcommand{\bp}{\mathbf{p}}
 \newcommand{\bc}{\mathbf{c}}
 \newcommand{\bft}{\mathbf{f}}
 \newcommand{\bz}{\mathbf{z}}
 \newcommand{\bv}{\mathbf{v}}
  \newcommand{\br}{\mathbf{r}}
 \newcommand{\bV}{\mathbf{V}}
 \newcommand{\bR}{\mathbb{R}}

 \newcommand{\mL}{\mathcal{L}}
 \newcommand{\mR}{\mathcal{R}}
 \newcommand{\plh}{\mkern-1.5mu{\times}\mkern-2mu}
 
 \newcommand{\wt}[1]{\widetilde{#1}}
 
 \newcommand{\ttt}[1]{\small\texttt{#1}}

\maketitle
\renewcommand{\thefootnote}{\roman{footnote}}

\begin{abstract}
   %
   %
   Photorealistic rendering of dynamic humans is an important ability for telepresence systems, virtual shopping, synthetic data generation, and more.
   %
   Recently, neural rendering methods, which combine techniques from computer graphics and machine learning, have created high-fidelity models of humans and objects.
   Some of these methods do not produce results with high-enough fidelity for driveable human models (Neural Volumes) whereas others have extremely long rendering times (NeRF).
   %
   %
   We propose a novel compositional 3D representation that combines the best of previous methods to produce both higher-resolution and faster results.
   Our representation bridges the gap between discrete and continuous volumetric representations by combining a coarse 3D-structure-aware grid of animation codes with a continuous learned scene function that maps every position and its corresponding local animation code to its view-dependent emitted radiance and local volume density.
   Differentiable volume rendering is employed to compute photo-realistic novel views of the human head and upper body as well as to train our novel representation end-to-end using only 2D supervision.
   In addition, we show that the learned dynamic radiance field can be used to synthesize novel unseen expressions based on a global animation code.
   Our approach achieves state-of-the-art results for synthesizing novel views of dynamic human heads and the upper body.
\end{abstract}

\section{Introduction}
Modeling, rendering, and animating dynamic human heads at high fidelity, for example for virtual reality remote communication applications, is a highly challenging research problem.
The main reason for this is the tremendous complexity of the human head in terms of geometry and appearance variations, e.g., of human skin, hair, teeth, and the eyes.
Skin exhibits subsurface scattering and shows fine-scale geometric pore-level detail, while the human eyes and teeth are both translucent and reflective at the same time.
High fidelity modeling and rendering of human hair is challenging due to its thin geometric structure and light scattering properties.
Importantly, the face is not static, but changes dynamically with expression and posture.

Recent work on neural rendering proposes to learn either discrete or continuous neural scene representations to achieve viewpoint and animation controllable rendering.
Discrete neural scene representations are based on meshes~\cite{thies2019neuraltex, liu2019softras, tran2019nonlinear3dmm, genova2018unsup3dmm, tewari2017self3dmm}, point clouds~\cite{wiles2020synsin, aliev2019neuralpoint, meshry2019neuralrerendering}, voxel grids~\cite{nvs_steve, sitzmann2019deepvoxels}, or multi-plane images~\cite{zhou2018mpi, mildenhall2019llff}.
However, each of these representations has drawbacks:
Meshes, even if dynamically textured~\cite{CodecAvatars}, struggle to model thin and detailed structures, such as hair.
Point clouds, by design, do not provide connectivity information and thus lead to undefined signals in areas of sparse sampling, while making explicit occlusion reasoning challenging.
Multi-plane images yield photo-realistic rendering results under constrained camera motion, but produce 'stack of cards'-like artifacts~\cite{zhou2018mpi} when the camera moves freely.
Volumetric representations~\cite{nvs_steve} based on discrete uniform voxel grids are
capable of modeling thin structures, e.g., hair, using semi-transparency.
While these approaches achieve impressive results, they are hard to scale up due to their innate cubic memory complexity.

To circumvent the cubic memory complexity of these approaches, researchers have proposed continuous volumetric scene representations based on fully-connected networks that map world coordinates to a local feature representation.
Scene Representation Networks (SRNs)~\cite{sitzmann2019srn} employ sphere marching to extract the local feature vector for every point on the surface, before mapping to pixel colors.
While this paper showed very inspiring results, the approach is limited to modeling diffuse objects, which makes it unsuitable to represent human heads at high fidelity.

Neural radiance fields~\cite{mildenhall2020nerf} have shown impressive results for synthesizing novel views of static scenes at impressive accuracy by mapping world coordinates to view-dependent emitted radiance and local volume density.
A very recent extension~\cite{liu2020nsvf} speeds up rendering by applying a static Octree to cull free space.
While they have shown first results on a simple synthetic dynamic sequence, it is unclear how to extend the approach to learn and render photo-realistic dynamic sequences of real humans.
In addition, it is unclear how to handle expression interpolation and the synthesis of novel unseen motions given the static nature of the Octree acceleration structure.

As discussed, existing work on continuous neural 3D scene representations mainly focuses on static scenes, which makes dynamic scene modeling and editing not directly achievable under the current frameworks.
In this work, we propose a novel compositional 3D scene representation for learning high-quality dynamic neural radiance fields that addresses these challenges.
To this end, we bridge the gap between discrete and continuous volumetric representations by combining a coarse 3D-structure-aware grid of animation codes with a continuous learned scene function. 
We start by extracting a global animation code from a set of input images using a convolutional encoder network.
The global code is then mapped to a 3D-structure-aware grid of local animation codes as well as a coarse opacity field.
A novel importance sampling approach employs the regressed coarse opacity to speed up rendering.
To facilitate generalization across motion and shape/appearance variation, in addition to conditioning the dynamic radiance field on the global animation code, we additionally condition it on a local code which is sampled from the 3D-structure-aware grid of animation codes.
The final pixel color is computed by volume rendering.
In summary, the main contributions of our work are:
%
\begin{itemize}
    \item A novel compositional 3D representation for learning high-quality dynamic neural radiance fields of human heads in motion based on a 3D-structure-aware grid of local animation codes.
    \item An importance sampling strategy tailored to human heads that allows to remove unnecessary computation in free space and enables faster volumetric rendering.
    \item State-of-the-art results for synthesizing novel views of dynamic human heads that outperform competing methods in terms of quality.
\end{itemize}


\section{Related Work}
Recently, there have been many works that combine deep neural networks with geometric representations to perform rendering.
In this section, we discuss different methods and their trade-offs, categorized by their underlying geometric representation.

\medskip\noindent\textbf{Mesh-based Representations:}
Triangle meshes have been used for decades in computer graphics since they provide an explicit representation of a 2D surface embedded within a 3D space.
A primary benefit of this representation is the ability to use high-resolution 2D texture maps to model high-frequency detail on flat surfaces.
Recently, differentiable rasterization~\cite{kato2018neurmeshrender, liu2019softras, chen2019nvdr, genova2018unsup3dmm, loper2014opendr, tran2019nonlinear3dmm, tewari2017self3dmm} has made it possible to jointly optimize mesh vertices and texture using gradient descent based on a 2D photometric re-rendering loss.
Unfortunately, these methods often require a good initialization of the mesh vertices or strong regularization on the 3D shape to enable convergence.
Moreover, these methods require a template mesh with fixed topology which is difficult to acquire.

\medskip\noindent\textbf{Point Cloud-based Representations:}
Point clouds are an explicit geometric representation that lacks connectivity between points, alleviating the requirement of a fixed topology but losing the benefits of 2D texture maps for appearance modeling.
Recent works, like
\cite{meshry2019neuralrerendering} and \cite{aliev2019neuralpoint}, propose methods that generate photo-realistic renderings using an image-to-image translation pipeline that takes as input a deferred shading deep buffer consisting of depth, color, and semantic labels.
Similarly, in SynSin~\cite{wiles2020synsin}, per-pixel features from a source image are lifted to 3D to form a point cloud which is later projected to a target view to perform novel view synthesis.
Although point clouds are a light-weight and flexible geometric scene representation, rendering novel views using point clouds results in holes due to their inherent sparseness, and it typically requires image-based rendering techniques for in-painting and refinement.

\medskip\noindent\textbf{Multi-plane Image-based Representations:}
Another line of work is using multi-plane images (MPIs) as the scene representation.
MPIs~\cite{zhou2018mpi} are a method to store color and alpha information at a discrete set of depth planes for novel view synthesis, but they only support a restricted range of motion.
LLFF~\cite{mildenhall2019llff} seeks to enlarge the range of camera motion by fusing a collection of MPIs~\cite{zhou2018mpi}.
Multi-sphere images (MSIs) \cite{attal2020msi,BroxtonVR} are an extension for the use case of stereo $360^\circ$ imagery in VR, where the camera is located close to the center of a set of concentric spheres.

\medskip\noindent\textbf{Voxel-based Representations:}
One big advantage of voxel-based representations is that they do not require pre-computation of scene geometry and that they are easy to optimize with gradient-based optimization techniques.
Many recent works~\cite{kar2017lmvsm, tulsiani2017multi, choy20163dr2n2, wu2016learningvox} have proposed to learn volumetric scene representation based on dense uniform grids.
Recently, such volumetric representations have attracted a lot of attention for novel view synthesis.
DeepVoxels~\cite{sitzmann2019deepvoxels} learns a persistent 3D feature volume for view synthesis with an image-based neural renderer.
Neural Volumes~\cite{nvs_steve} proposes a differentiable raymarching algorithm for optimizing a volume, where each voxel contains an RGB and transparency values.
The main challenge for voxel-based techniques originates in the cubic memory complexity
of the often employed dense uniform voxel grid, which makes it hard to scale these approaches
to higher resolutions.

\medskip\noindent\textbf{Implicit Geometry Representations:}
Implicit geometry representations have drawn a lot of attention from the research community due to their low storage requirements and the ability to provide high-quality reconstructions with good generalization.
This trend started with geometric reconstruction approaches that first employed learned functions to represent signed distance fields (SDFs)~\cite{park2019deepsdf, jiang2020localsdf, chabra2020deeplocalshape} or occupancy fields~\cite{Mescheder2019occnet, genova2020localsdf, Peng2020convoccnet}.
DeepSDF~\cite{park2019deepsdf} and OccNet~\cite{Mescheder2019occnet} are among the earliest works that try to learn an implicit function of a scene with an MLP and are fueled by large scale 3D shape datasets, such as ShapeNet~\cite{chang2015shapenet}.
DeepSDF densely samples points around the surface to create direct supervision for learning the continuous SDF, while OccNet learns a continuous occupancy field.
ConvOccNet~\cite{Peng2020convoccnet} manages to improve OccNet's ability to fit large scale scenes by introducing a convolutional encoder-decoder.
ConvOccNet is limited to static scenes and geometry modeling, i.e., it can not handle the dynamic photo-realistic sequences that are addressed by our approach.
%
%

\medskip\noindent\textbf{Continuous Scene Representations:}
Inspired by their implicit geometry counterparts, continuous scene representations for modeling colored scenes have been proposed.
Scene Representation Networks (SRNs)~\cite{sitzmann2019srn} propose an approach to model colored objects by training a continuous feature function against a set of multi-view images.
DVR~\cite{Niemeyer2020DVR} derived an analytical solution for the depth gradient to learn an occupancy and texture field from RGB images with implicit differentiation.
NeRF~\cite{mildenhall2020nerf} learns a 5D neural radiance field using differentiable raymarching by computing an integral along each ray.
Although promising, their results are limited to a single static scene and the approach is hard to generalize to multiple scenes or a scene with dynamic objects.
%
%
Another limiting factor is that these representations are extremely costly to render, since every step along the ray requires an expensive evaluation of the complete fully-connected network.
GRAF~\cite{schwarz2020graf} introduces a generative model for radiance fields which extends NeRF's ability to model multiple static objects.
However, their approach is limited to very simple scenes at low resolution.
The approach has not been demonstrated to scale to the dynamic high-quality real-world animations we are interested in.
A very recent work, NSVF~\cite{liu2020nsvf}, manages to solve the second limitation with an Octree acceleration structure.
%
Although they provide initial results on a single synthetic dynamic sequence, the static Octree structure is optimized frame by frame rather than regressed from temporal information, which makes it not straightforward to directly deploy their method on novel photo-realistic dynamic sequences of real humans.
In addition, it is unclear how to efficiently handle expression interpolation and novel motion synthesis given their static Octree.

\section{Method}
\begin{figure*}[!ht]
\begin{center}
\includegraphics[width=0.95\textwidth]{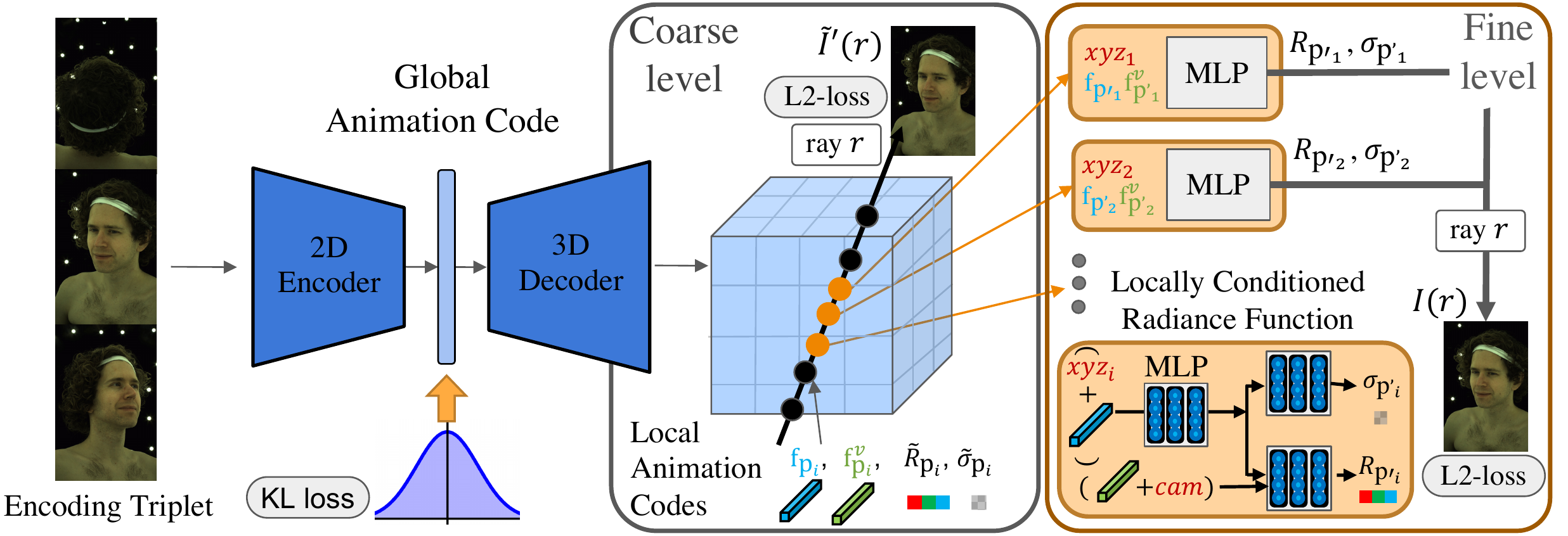}
\end{center}
   \vspace{-0.3cm}
   \caption{\textbf{Method overview}. Given a multi-view video as input, we learn a dynamic radiance field parametrized by a global animation code. To render a particular frame, the global code is first mapped to a coarse voxelized field of local animation codes using a 3D convolutional decoder. This grid of animation codes provides local conditioning at each 3D position for the fine-level radiance function, represented as an MLP. Differentiable ray marching is used to render images and provide supervision, and can be sped up significantly by using a ray sampling strategy that
   uses the coarse grid to determine relevant spatial regions.
   }
\label{fig:teaser1}
\end{figure*}

In this section, we introduce our novel compositional representation 
that combines the modeling power of high-capacity voxel-based
representations and the ability of continuous scene representations 
to capture subtle fine-level detail. 
In Fig.~\ref{fig:teaser1} we provide an overview of our approach.

The core of the method is a hybrid encoder-decoder architecture, 
directly supervised with multi-view video sequences. 
For a given frame, the encoder takes a sparse set of views,
and outputs a global animation code, which describes dynamic scene 
information specific to the frame. 
The global animation code is used to condition the 3D convolutional decoder, 
which outputs a coarse 3D structure-aware voxel field.
In particular, each voxel stores coarse-level opacity, color and 
localized animation codes, which represent local dynamical properties of the 
corresponding spatial region of the scene.
The resulting voxel field is further used to create a coarse volumetric rendering
of the scene, which may lack fine-level detail, but provides a reliable 
initial estimate of the scene's geometry, which is crucial to enable efficient
continuous scene modeling.
To account for the lack of detail, we rely on a continuous scene function, 
represented as an MLP, to model fine-level radiance. 
The coarse-level geometry estimate is used to define spatial regions where 
the function is evaluated, and the local animation codes as spatially-varying 
conditioning signal to the MLP.
To better model view-dependent effects, both coarse- and fine-level 
representations are partly conditioned on the camera viewpoint.
The outputs of the continuous scene function are then used to create the 
final, refined volumetric rendering of the scene. 
In what follows, we describe each of the components in detail.

\subsection{Encoder-Decoder}

The goal of the encoder is to produce a compact representation that 
captures global dynamical properties of the scene, which then serves as a
conditioning signal for the decoder.
Our encoder is a 2D convolutional network which takes a sparse set 
of views and outputs parameters of a diagonal Gaussian distribution 
$\bm{\mu}, \bm{\sigma} \in \bR^{256}$.
In practice, the encoder is conditioned on three different camera views, concatenated
along the channel axis. Given the distribution $\mathcal{N}(\bm{\mu}, \bm{\sigma})$,
we use the reparameterization trick to produce the global animation
code $\bz \in \bR^{256}$ in
a differentiable way, and pass it to the decoder. 
We found that using a variational formulation~\cite{kingma2013vae}
is critical for making our model animatable.

Given the global animation code $\bz$, the goal of the decoder is to 
produce a coarse-level representation of the scene. 
In particular, the coarse level is modeled by a volumetric field
\vspace{-0.1cm}
\begin{equation}
    \mathbf{V}_{\bp} = \left( 
    \wt{\bc}_{\bp}, \wt{\sigma}_{\bp}, \bft_{\bp}, \bft^{v}_{\bp}   \right) \; ,
\end{equation}
where $\wt{\bc}_{\bp} \in \bR^3$ is a coarse-level color value, 
$\wt{\sigma}_{\bp} \in \bR$ is differential opacity,
$\bft_{\bp} \in \bR^{32}$ is the view-independent local animation code,
$\bft_{\bp}^v \in \bR^{32}$ is the view-dependent local animation code,
and $\bp \in \bR^3$ is the spatial location. In our framework, $\mathbf{V}$ 
is produced by a volumetric decoder as an explicit coarse discrete grid 
$\mathbf{G} \in \bR^{D\plh D \plh D \plh F}$, where $D = 64$ is
the spatial dimension of the grid, and $F = 68$ is the dimensionality of the field. 
Samples $\bV_{\bp}$ at continuous locations $\bp \in \bR^3$ are produced
with trilinear interpolation over the voxels.

In practice, the decoder is represented by two independent 3D convolutional
neural network branches. 
The first branch is conditioned only on
the global code $\bz$, and predicts view-independent values, 
the differential occupancy $\wt{\sigma}_{\bp}$ 
and the view-independent local animation codes $\bft_{\bp}$.
The second branch predicts view-dependent color values $\wt{\bc}_{\bp}$ and
local animation codes $\bft_{\bp}^v$, and is conditioned on both
the global code $\bz$ and the viewpoint $\bv \in \bR^3$,
which is computed as a normalized difference between the camera location 
and the center of the scene.

\subsection{Volumetric Rendering}
\label{sec:volumetric-rendering}

Given the discrete voxel field, we apply differentiable 
ray-marching to obtain coarse volumetric rendering~\cite{mildenhall2020nerf}. 
Namely, for each ray $\mathbf{r} \in \bR^3$ shot from the camera 
center $\mathbf{o} \in \bR^3$, we sample $N$ query points
$\bp_i = (\mathbf{o}+d_i \cdot \mathbf{r})$ along $\br$, 
where $d_i$ is the depth sampled uniformly between the depth at a near plane
$d_{min}$ and a far plane $d_{max}$.
Estimates of expected coarse opacity $\wt{A}_{\br}$ and 
color $\wt{I}'_{\br}$ are then computed as 
\vspace{-0.1cm}
\begin{align}
    \wt{A}_{\br} &= \sum_{i=1}^{N} T_i \alpha_i \; , &
    \wt{I}'_{\br} &= \sum_{i=1}^{N} T_i \alpha_i \wt{\bc}_{\bp_i} \;, 
    \label{eq:coarse-accumulate}
\end{align}
where $T_i = \exp(-\sum_{j=1}^{i-1}\wt{\sigma}_{\bp_j}\delta_j)$, $\alpha_i = (1-\exp(-\wt{\sigma}_{\bp_i} \delta_i))$, 
and $\delta_i = \|d_{i+1}-d_{i}\|$ is the distance between
two neighbouring depth samples. In practice, values $\wt{\bc}_{\bp_i}$, 
$\wt{\sigma}_{\bp_i}$ are sampled from the voxel grid with trilinear interpolation.

The final coarse-level rendering is computed by compositing
the accumulated color $\wt{I}'_{\br}$ and the background color
with a weighted sum
\vspace{-0.1cm}
\begin{equation}
    \wt{I}_{\br} = \wt{I}'_{\br} + (1 - \wt{A}_{\br}) I^{bg}_{\br} \; .
    \label{eq:coarse-blending}
\end{equation}
The resulting coarse rendering roughly captures the appearance of the scene,
but lacks fine-level detail. A seemingly straightforward way 
to improve the level of detail would be to increase the spatial resolution of 
the voxel grid. Unfortunately, this quickly becomes impractical due 
to the cubic memory complexity of these representations.

\subsection{Continuous Scene Function}

In order to improve fine-level modeling capabilities while avoiding
heavy memory costs associated with high-res voxel representations, 
we introduce a \textit{continuous} scene function $f(\cdot)$, 
parameterized as an MLP.
The key intuition is that voxel-based approaches represent scenes 
\textit{explicitly} and uniformly across space, thus often wasting resources on 
irrelevant areas. On the other hand, continuous representations are 
\textit{implicit}, and allow for more flexibility, as the scene function can 
be evaluated at arbitrary locations. When combined with a sampling strategy
that focuses only on relevant spatial locations, this flexibility can 
bring significant efficiency improvements.

One crucial difference of our method with respect to the existing continuous
neural rendering approaches~\cite{schwarz2020graf, mildenhall2020nerf}, is 
that in addition to conditioning on the location, view direction and the
global scene information, our scene function is also conditioned on 
spatially-varying local animation codes. 
As we demonstrate in our experiments in Sec.~\ref{sec:experiments}, 
this increases the effective capacity of our model, and allows our model
to capture significantly more detail and better generalize 
across different motion and shape/appearance variations. We also show
that this is especially important for modeling dynamic scenes, as
they require significantly more modeling capacity and the naive MLP-based
approaches typically fail.

More formally, the scene function $f(\cdot)$ takes as inputs coordinates of a 
sampled query point $\mathbf{p}$, view vector $\mathbf{v}$, and 
the corresponding local animation codes $\bft_{\bp}, \bft^v_{\bp}$,
and produces the fine-level color $\bc_{\bp} \in \bR^3$ and the differential 
probability of opacity $\sigma_{\bp} \in \bR$
\vspace{-0.1cm}
\begin{equation*}
\bc_{\bp}, \sigma_{\bp} = f(\phi(\bp),\phi(\bv),\bft_{\bp},\bft^v_{\bp}) \;.
\end{equation*}
Feature vectors $\bft_{\bp},\bft^v_{\bp}$ are obtained from the 
the coarse voxel grid via trilinear interpolation,
and position $\bp$ and view $\bv$ vectors are passed through 
a positional encoding $\phi(\cdot)$, in order to better
capture high-frequency information~\cite{mildenhall2020nerf}.

Fine-level rendering $I_{\br}$ and $A_{\br}$ can then be computed by evaluating $f(\cdot)$
at a number of sampled query points along each ray and applying Eq.~(\ref{eq:coarse-accumulate})-(\ref{eq:coarse-blending}). 
In the next section, we discuss our novel sampling scheme 
that allows to significantly speed up the rendering process. 

\subsection{Efficient Sampling}
Using spatially-varying conditioning allows us to increase effective
capacity of our continuous scene representation and leads to better
generalization. However, producing a high-quality rendering still
requires evaluating the scene function at a large number of query
locations, which can be computationally expensive~\cite{mildenhall2020nerf}, 
and ultimately suffers from similar limitations as the voxel fields.
Luckily, we can exploit the fact that our coarse voxel field
already contains information about the scene's geometry. To this end, 
we introduce a simple and efficient sampling scheme, which uses the coarse 
opacity values to produce a strong initial prior on the underlying geometry. 
In particular, for each ray $\br$, we first compute a coarse 
depth $\wt{d}_{\br}$ as
\vspace{-0.1cm}
\begin{align*}
    \wt{d}_{\br} &= \frac{1}{\wt{A}_{\br}}
    \sum_{i=1}^{N} T_i \alpha_i \cdot d_i \; ,
\end{align*}
where $d_i$ are the \textit{same} uniform samples as in Eq.~(\ref{eq:coarse-accumulate}).
Then, we obtain our new fine-level location samples from a uniform distribution:
\vspace{-0.1cm}
\begin{equation*}
 d \sim \mathcal{U}\left[\wt{d}_{\br} - \Delta_d, \wt{d}_{\br} + \Delta_d \right] \; ,
\end{equation*}
centered at the depth estimate $\wt{d}_{\br}$, where $\Delta_d = \frac{(d_{max}-d_{min})}{k}$,
i.e. $k = 10$ times smaller range than at the coarse level.
In Sec.~\ref{sec:experiments} we demonstrate that this strategy in practice
leads to comparable rendering quality, while being more computationally
efficient.

\subsection{Training Objective}
Our model is end-to-end differentiable, which allows us to jointly train 
our encoder, decoder and the scene MLP, by minimizing the following loss:
\vspace{-0.1cm}
\begin{equation*}
    \mL = \mL_{r} + \wt{\mL}_{r} + \lambda_f{\mL}_{\beta} + \lambda_c\wt{\mL}_{\beta} + \lambda_{\mathrm{KL}}\mL_{\mathrm{KL}} \;.
\end{equation*}
Here $\mL_{r}$ is the error between the rendered and ground truth images for the fine-level
rendering:
\vspace{-0.1cm}
\begin{align*}
    \mL_{r} &= \sum_{\br \in \mathcal{R}}
    ||I_{\br} - I^{gt}_{\br}||_2^2 \;,
\end{align*}
where $\mathcal{R}$ is a set of rays sampled in a batch. 
The coarse-level rendering loss $\wt{\mL}_r$ is computed similarly. 
${\mL}_{\beta}$ and $\wt{\mL}_{\beta}$ are the priors on the fine-level and coarse-level image opacities respectively~\cite{nvs_steve}:
\vspace{-0.1cm}
\begin{align*}
\mL_\beta = \sum_{\br \in \mathcal{R}}  \left( \log {A}_{\br} + \log (1 - {A}_{\br})  \right) \;,
\end{align*}
which pushes both the coarse and fine opacities to be sharper, and encodes the prior belief 
that most of the rays should hit either the object or the background. Finally, 
the Kullback-Leibler divergence loss $\mathcal{L}_{\mathrm{KL}}$ encourages
our global latent space to be smooth~\cite{kingma2013vae},
which improves the animation and interpolation capabilities
of our model.

\section{Experiments}
\label{sec:experiments}
We first compare with two state-of-the-art methods for novel view synthesis, namely NV~\cite{nvs_steve} and NeRF~\cite{mildenhall2020nerf} on four dynamic sequences of a human head making different facial expressions or talking.
We then perform an ablation study to test how different feature representations affect the ability to capture longer sequences, as well as the effects of applying different resampling strategies on speed and image quality.
We also evaluate generalization capabilities of our model on novel sequence 
generation and animation, by interpolating in latent space and by driving the model with 
various input modalities, including keypoints and images.

\subsection{Datasets}
%
We use a multi-camera system with around 100 synchronized color cameras that produces $2048\times1334$ resolution images at 30 Hz. The cameras are distributed approximately spherically at a distance of one meter, and focused at the center of the capture system to provide as many viewpoints as possible. Camera intrinsics and extrinsics are calibrated in an offline process.
Images are downsampled to $1024\times667$ for training and testing.
Each capture contains $n=3$ sentences and around $k=350$ frames in total for each camera 
view.
We trained on ~$m=93$ cameras and tested on $q=33$ frames from another $p=7$ cameras.

\subsection{Baselines}
We compare our methods with two baselines that we describe in the following. 

\textbf{NV~\cite{nvs_steve}}:
Neural Volumes performs novel view synthesis of a dynamic object-centric scene by doing raymarching on a warped voxel grid of RGB and differential opacity that is regressed from three images using an encoder-decoder network.
As the volume is conditioned on temporal input of RGB images, NV is capable of rendering dynamic scenes.
The volume is of size $128^3$ and the warp field is $32^3$.
The global animation code is a feature vector of $256$ entries.

\textbf{NeRF~\cite{mildenhall2020nerf}}:
NeRF learns a continuous function of scene radiance, including RGB and opacity, with a fully connected neural network conditioned on scene coordinates and viewing direction.
Positional encoding is applied to the 3D coordinates to better capture high frequency information, and raymarching is performed to render novel views.
Note that the original NeRF approach is not directly applicable to dynamic sequences.
Thus, we extend the conditioning signal to NeRF with a global animation code generated from 
the encoder in NV.
The global animation code is generated from the encoder in NV and it is also of size $256$.

\subsection{Novel View Synthesis}
We show quantitative and qualitative results of novel view synthesis on four dynamic sequences of human heads.

\noindent\textbf{Quantitative Results}:
We report quantitative evaluation results in Tab.~\ref{tab:nvs1}.
Metrics used here are MSE, PSNR, and SSIM.
We average those metrics across different test views and time steps, among each of the sequences.
%
%
%
To compensate for sensory difference between each camera, we apply the same color calibration network as in NV~\cite{nvs_steve} for our methods as well as all baselines.
To compute the parameters of the color calibration networks, we first fit the color calibration model on an additional sentence with all camera views and fix the parameters for all subsequent steps.
%
The first three sequences (Seq1-Seq3) are captures showing the participant talking, while the last one (Seq4) is a capture of a range of motions showing challenging expressions.
As we can see, our method outperforms all other baselines on the four dynamic sequences in terms of all metrics.

\begin{table*}[]
\centering
\begin{tabular}{l|lll|lll|lll|lll|}
\multirow{2}{*}{} &
  \multicolumn{3}{c|}{Sequence1} &
  \multicolumn{3}{c|}{Sequence2} &
  \multicolumn{3}{c|}{Sequence3} &
  \multicolumn{3}{c|}{Sequence4} \\ \cline{2-13} 
 &
  \multicolumn{1}{l|}{MSE} &
  \multicolumn{1}{l|}{PSNR} &
  SSIM &
  \multicolumn{1}{l|}{MSE} &
  \multicolumn{1}{l|}{PSNR} &
  SSIM &
  \multicolumn{1}{l|}{MSE} &
  \multicolumn{1}{l|}{PSNR} &
  SSIM &
  \multicolumn{1}{l|}{MSE} &
  \multicolumn{1}{l|}{PSNR} &
  SSIM \\ \hline
NV  & 46.19 & 31.56 & 0.8851 & 52.11 & 31.24 & 0.8499 & 83.07 & 29.24 & 0.7742 & 40.47 & 32.30 & 0.9086 \\
NeRF & 43.34 & 31.88 & 0.8923 & 46.89 & 31.79 & 0.8531 & 90.45 & 28.87 & 0.7727 & 35.52 & 32.95 & 0.9129 \\
Ours & \textbf{34.01} & \textbf{33.09} & \textbf{0.9064} & \textbf{42.65} & \textbf{32.24} & \textbf{0.8617} & \textbf{79.29} & \textbf{29.61} & \textbf{0.7826} & \textbf{27.62} & \textbf{34.12} & \textbf{0.9246}
\end{tabular}
\caption{\label{tab:nvs1}\textbf{Image prediction error}. We compare NV, NeRF, and our method on 4 sequences, and report average error computed over a set of approximately 200 images of 7 views for each sequence.
Our method outperforms all other baselines on all metrics.
}
\end{table*}

\begin{figure}
    \centering
    \includegraphics[width=0.49\textwidth]{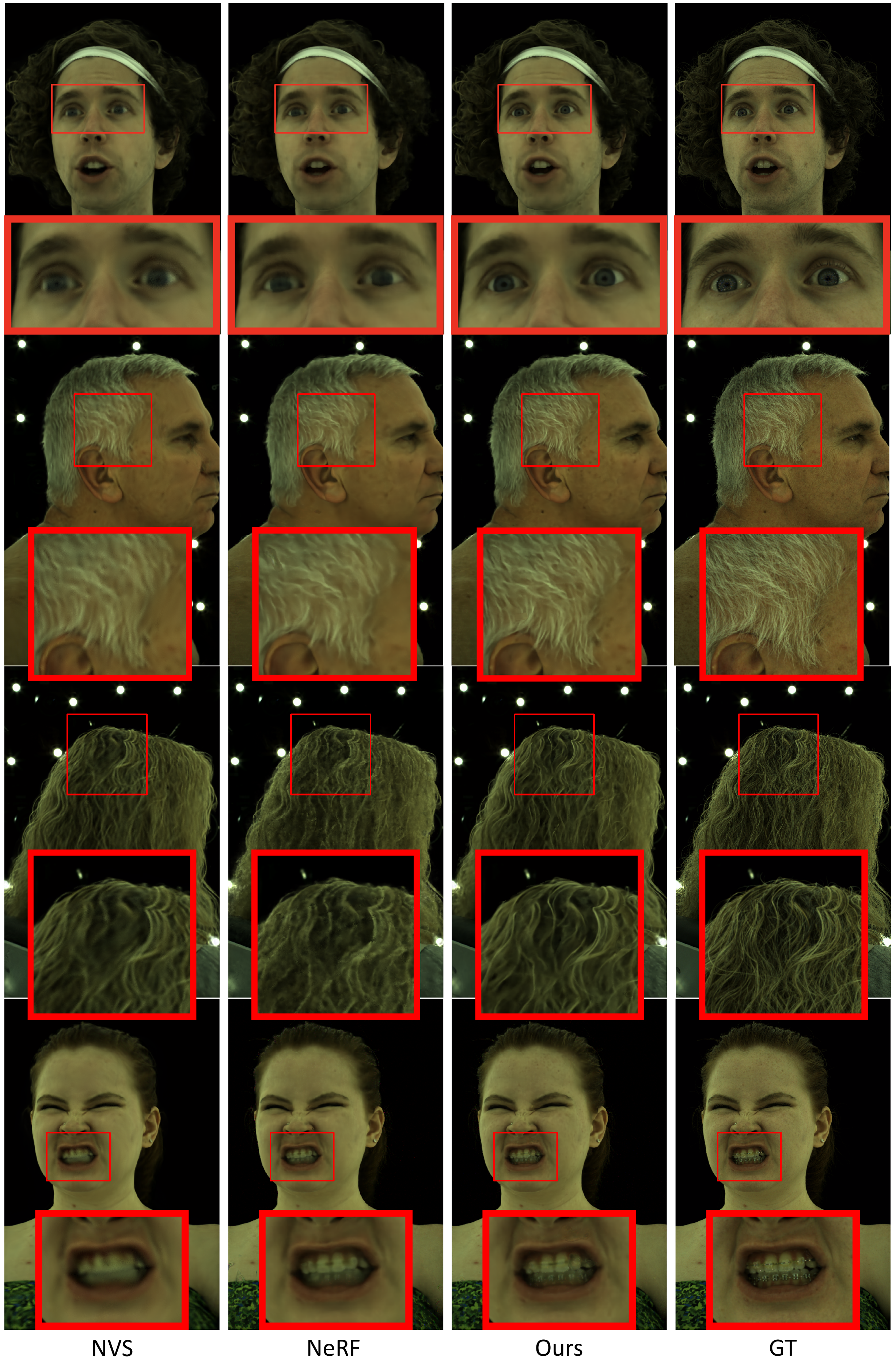}
    \caption{
    \textbf{Qualitative comparison of rendered images}.
    Our method recovers more fine-scale details than NV and NeRF, particularly in high-frequency regions like the eyes and hair.
    Results are rendered at $1024\times667$ with insets for better visualization.
    } 
    \label{fig:vis1}
\end{figure}

\noindent\textbf{Qualitative Results}:
We show visual comparisons between different models trained on long video sequences in Fig.~\ref{fig:vis1}.
We can see that NV and NeRF trained on a sequence tend to yield relatively blurry results, while our approach produces sharper images and can reconstruct finer details in terms of both texture and geometry on areas like hair, eyes, and teeth. Our method can achieve photo-realistic rendering results on video sequences.



\subsection{Ablation Studies}
\begin{figure}
\centering
\includegraphics[width=0.45\textwidth]{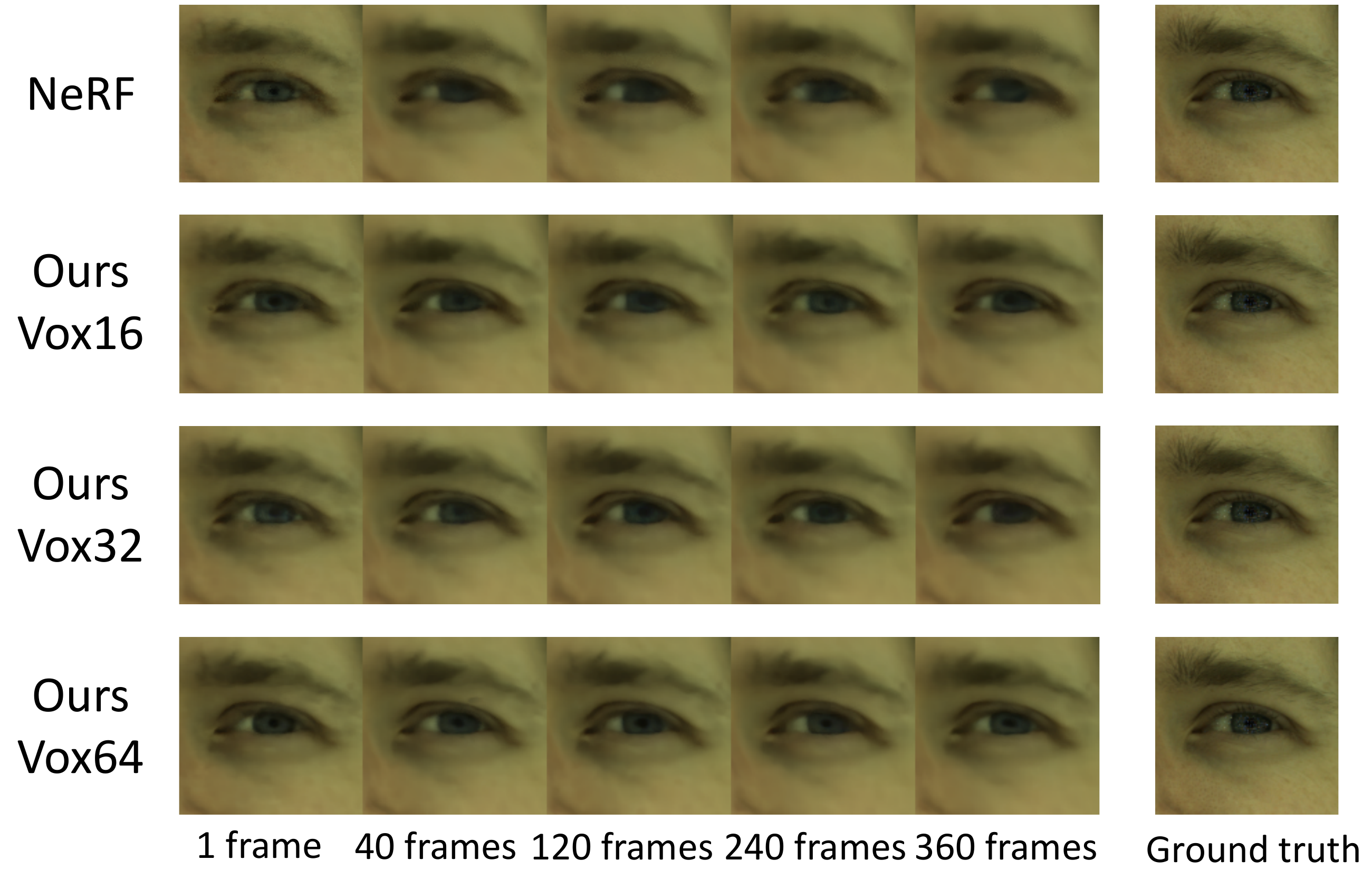}
\caption{
\textbf{Effect of sequence length on quality.}
Conditioning the radiance field using local animation codes instead of a global code greatly expands model capacity, allowing our model to recover much sharper images even when trained on longer video sequences.
}
\label{fig:abla1}
\end{figure}

\begin{figure}
\centering
\includegraphics[width=0.5\textwidth]{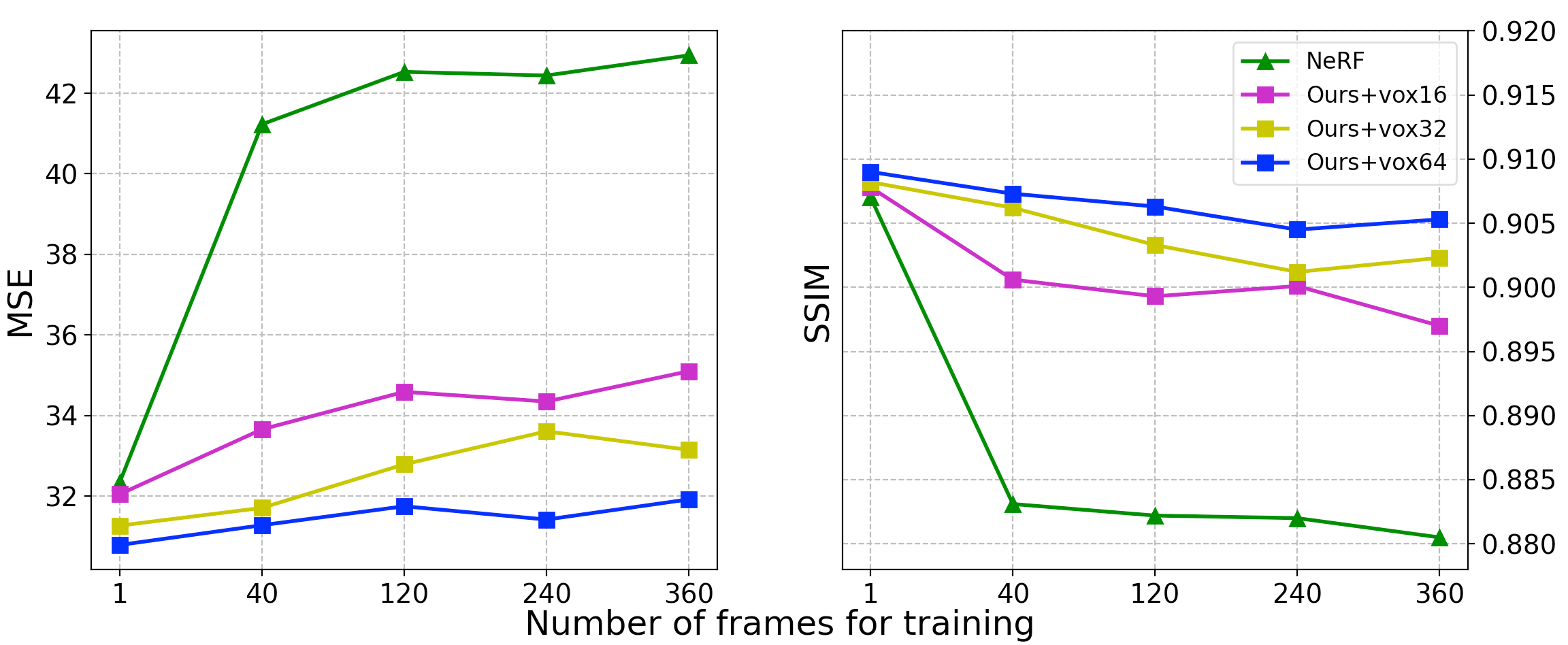}
\caption{
\textbf{Effect of sequence length on reconstruction.} MSE and SSIM on the first frame v.s.~length of the training sequence.
}
\label{fig:seq_len_mse}
\end{figure}

\noindent\textbf{Longer Sequences}:
%
%
As one of the major differences between our method and the adapted NeRF is the different feature representation as input for the fine-level neural implicit function, we also tested how this impacts the generalization and fitting power of the approaches.
To achieve that, we train our method as well as the temporal conditioned NeRF on sequences with variable length (1, 40, 120, 240, 360 frames) and report their reconstruction performance on a set of views at certain time frames.
For all training sets, the first frame is shared and is taken as the test frame.
For comparisons, we evaluate three different resolutions (16, 32, 64) for the coarse-level voxel feature in our method to better understand how the voxel resolution could affect the generalization capabilities and expressiveness of our model. 
Figure~\ref{fig:seq_len_mse} shows the plot of MSE and SSIM v.s.~the length of the training sequence of different models.
A direct visual comparison between models trained on a different number of frames is shown in Figure~\ref{fig:abla1}.
As can be seen, the performance of NeRF with a global animation code drops significantly when the total number of training frames increases, while our method maintains a higher rendering quality due to a more expressive local voxel animation code, which enforces the fine-level implicit function to learn local rather than global representations and capture high frequency details more accurately.
In addition, the 3D convolutional decoder imposes 3D inductive bias and improves the capacity of the whole model with the help of a 3D voxel feature tensor that has more spatial awareness compared to a global code.
%
We also see that rendering quality improves and the model achieves better generalization when the coarse-level voxel feature resolution is relatively large.
As can be seen in Fig.~\ref{fig:abla1}, when the resolution is smaller, the performance drops as each local code is responsible for describing a larger region of space.


\medskip\noindent\textbf{Sampling Strategy and Runtime Comparison}:
We further trained our method and NeRF on a single sentence applying different sampling schemes: the hierarchical sampling (HS) in~\cite{mildenhall2020nerf} and our simple sampling (SS). We show results in Tab.~\ref{tab:abla_sample}. As we can see our simple sampling preserves rendering quality while enjoying a large increase in runtime efficiency.
For rendering an image with resolution $1024\times667$, NV takes roughly $0.9$s and NeRF is taking $\textgreater{}25$s whereas our methods takes $3.6$s.

\begin{table}
\centering
\begin{tabular}{c|ccc|c}
        & \multicolumn{1}{c|}{MSE} & \multicolumn{1}{c|}{PSNR} & SSIM   & Runtime           \\ \hline
NeRF+HS & 36.33                    & 32.90                     & 0.8898 & \textgreater{}25s \\
NeRF+SS & 38.80                    & 32.75                     & 0.8886 & 19.69s              \\ \hline
Ours+HS & \textbf{27.23}                    & \textbf{34.24}                     & 0.9090 & 14.30s            \\
Ours+SS & 30.35                    & 34.13                     & \textbf{0.9113} & \textbf{3.6s}           
\end{tabular}
\caption{\label{tab:abla_sample}\textbf{Ablation on different sampling schemes.} We show image reconstruction results as well as runtime for both NeRF and ours with different sampling strategies.}
\end{table}

\subsection{Animation}
\begin{figure*}
    \centering
    \includegraphics[width=1.0\textwidth]{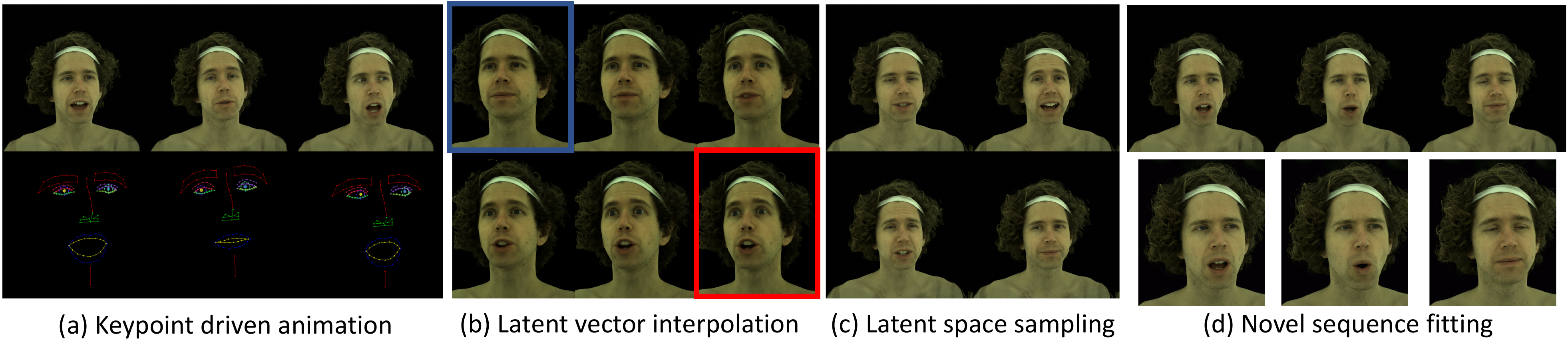}
    \caption{
    \textbf{Novel sequence generation}.
    New animations can be created by dynamically changing the global animation code, for example by (a) using keypoints to drive the animation, (b) interpolating the code at key frames, (c) sampling from the latent distribution, or (d) directly fitting the codes to match a novel sequence.
    Please refer to supplemental material for more visual results.
    }
    \label{fig:anim_all}
\end{figure*}
We demonstrate a large variety of applications that is enabled by our approach.

\medskip\noindent\textbf{Latent Space Sampling and Interpolation}
Given the encoder-decoder architecture, we can generate smooth transitions between two expressions by interpolating in latent space and create free-view animations.
In Fig.~\ref{fig:anim_all}(b), we show direct interpolation results between two frames with different expressions.
The frames in the red and blue bounding box are two key frames and all other frames inbetween are interpolated results.
We also show rendering results by randomly sampling in the latent space in Fig.~\ref{fig:anim_all}(c).
%


%

\begin{table}
\centering
\begin{tabular}{c|ccc}
\hline
                      & MSE   & PSNR  & SSIM   \\ \hline
keypoints encoder \textbf{w/o} ft     & 58.52 & 30.78 & 0.8891 \\
image encoder \textbf{w/o} ft    & 55.86 & 31.07 & 0.8903 \\ \hline
keypoints encoder \textbf{w/} ft & 35.12 & 32.90 & 0.9024 \\
image encoder \textbf{w/} ft & 34.86 & 33.27 & 0.9053 \\ \hline
full model ft & \textbf{32.47} & \textbf{33.84} & \textbf{0.9121} \\ \hline
\end{tabular}
\caption{
\textbf{Novel content synthesis.}
We show results on novel content generation and novel sequence fitting.
We tested two different encoder models that use data from two modalities: sparse 2D keypoints and images.
}
\label{tab:anim}
\end{table}

\medskip\noindent\textbf{Landmark Driven Animation}:
Because the decoder only depends on a single global code to generate the dynamic field, the original image encoder can be switched to an encoder that takes inputs from other modalities as long as correspondence between input and outputs can be established.
To demonstrate controllable animation, we use 2d landmarks as a substitute of the image input and train a simplified PointNet~\cite{qi2017pointnet}-like encoder that regresses the global code from the set of 2d landmarks.
To train such an encoder, we minimize the $\ell_2$ distance between the global code $z_{kps}$ from keypoints and its corresponding global code $z_{img}$ from the image on the training set.
Fig.~\ref{fig:anim_all}(a) shows some rendering results that are driven by a keypoint encoder.
%
%
To test generalization to a novel sentence that is not included in the training data, we deployed the keypoint encoder and the pretrained decoder on a novel sequence.
Results on test views are reported in Tab.~\ref{tab:anim}.
We can see, that with a keypoint encoder using only a regression loss in the latent space, the avatar can be driven with reasonable performance, even though keypoints provide less information than images.

\medskip\noindent\textbf{Novel Sequence Fitting}:
To demonstrate our model's ability to generalize to a novel sequence, we show results of animations driven by novel video sequences.
For novel sequence generation from a given input modality, two components need to generalize: (1) the encoder, which produces animation codes given novel image inputs, and (2) the decoder, which renders novel animation codes into images. 
We first study the generalization ability of the decoder in isolation. To do this, we fine-tune the encoder on the novel sequence, fixing the parameters of the decoder and only back-propagating gradients to the encoder's parameters.
Fig.~\ref{fig:anim_all}(d) shows rendering results.
%
%
To test the ability of generalization to novel input driving sequences, we test the complete encoder-decoder model on a novel sequence, without any fine-tuning.
Results are shown in Tab.~\ref{tab:anim}.
As we can see, an image-based encoder trained with a photometric loss shows better performance on novel content than a key-point encoder trained with a regression loss on the latent space.
Innately, image input is a more informative input than sparse key-points.
Training with a photometric loss rather than a regression loss enables the encoder to output latent codes that are more compatible with the decoder.
We also fine-tuned just the image encoder with a photometric loss to align the latent space and we find that the rendering results achieve compatible quality on novel content.
We also fine-tuned the full model (both encoder and decoder) and we find the gap is not large in comparison to the model that only has its encoder fine-tuned.
\section{Limitations}
While we achieve state-of-the-art results, our approach is still subject to a few 
limitations which can be addressed in follow-up work:
(1) Our method heavily relies on the quality of the coarse-level voxel field.
In cases when the voxel representation has significant errors, the 
following fine-level model is likely not to recover.
(2) Since we rely on the voxel field for our coarse-level representation, 
our method is primarily applicable to object-centric scenes.
Potentially, by substituting the voxelized representation with a coarse depth map, it could also be applied to arbitrary scenes.
(3) Although our compositional approach improves the scalability
of both voxel-based and continuous representations, our approach is 
still limited in terms of resolution.
One possibility to tackle this could be to also regress the positions and locations of the voxels or a group of voxels, which could serve as a more efficient and reliable proxy.

\section{Conclusion}

In this paper, we proposed a method for rendering and driving photo-realistic avatars of humans captured with a multi-view camera system.
Our representation bridges the gap between discrete and continuous volumetric representations by combining a coarse 3D-structure-aware grid of animation codes with a continuous learned scene function that enables high-resolution detail without the need for a dense voxel grid.
We show that our approach produces higher-quality results than previous methods, especially as the length of the sequence increases, and is significantly faster than classical neural radiance fields.
Our approach also enables driving the model, which we demonstrate via interpolation in the latent space, randomly sampling the latent space, and facial motion control via a set of sparse keypoints.
We believe that our approach is a stepping stone towards higher-quality telepresence systems.

%

{\small
\bibliographystyle{ieee_fullname}
\bibliography{egbib}
}

\end{document}


 \newcommand{\TB}[1]{{\color{green}{\bf TB: #1}}}
  
 \newcommand{\tb}[1]{{\color{green} #1}}
 
 \newcommand{\TS}[1]{{\color{BurntOrange}{\bf TS: #1}}}
 
 \newcommand{\ZW}[1]{{\color{blue}{\bf ZW: #1}}}
 
 \newcommand{\SL}[1]{{\color{gray}{\bf SL: #1}}}
 
 \definecolor{mzcolor}{RGB}{255,50,00}
 \newcommand\MZ[1] {\textbf{\textcolor{mzcolor}{MZ: #1}}}
 
 \newcommand{\bp}{\mathbf{p}}
 \newcommand{\bc}{\mathbf{c}}
 \newcommand{\bft}{\mathbf{f}}
 \newcommand{\bz}{\mathbf{z}}
 \newcommand{\bv}{\mathbf{v}}
  \newcommand{\br}{\mathbf{r}}
 \newcommand{\bV}{\mathbf{V}}
 \newcommand{\bR}{\mathbb{R}}

 \newcommand{\mL}{\mathcal{L}}
 \newcommand{\mR}{\mathcal{R}}
 \newcommand{\plh}{\mkern-1.5mu{\times}\mkern-2mu}
 
 \newcommand{\wt}[1]{\widetilde{#1}}
 
 \newcommand{\ttt}[1]{\small\texttt{#1}}

\title{
Learning Compositional Radiance Fields of Dynamic Human Heads\\
-- Supplemental Document --
}

\author{First Author\\
Institution1\\
Institution1 address\\
{\tt\small firstauthor@i1.org}
\and
Second Author\\
Institution2\\
First line of institution2 address\\
{\tt\small secondauthor@i2.org}
}

\maketitle
\thispagestyle{empty}

\appendix

\section{Video Results}
Please refer to the \texttt{videos} directory for video results and \texttt{index.html} for a video navigation.

\section{Training Details}
\subsection{Network Architecture}

There are three main neural networks used in our methods: 
1) \textbf{Encoder}, that regresses image input to the statistics $\bm{\mu}, \bm{\sigma}$ of a latent space vector $\bz \in \bR^{256}$;
2) \textbf{Decoder}, a 3D convolutional network that regresses the latent vector $\bz$ to a coarse-level volume $\mathbf{V}_{\bp}$ of log differential opacity $\wt{\sigma}_{\bp}$, color $\wt{\bc}_{\bp}$, and spatial scene features $\bft_{\bp}, \bft^{v}_{\bp}$;
3) \textbf{Refinement MLP}, that takes in the coordinate of a spatial location $\bp$ as well as its corresponding spatial local feature from the coarse-level volume $\bft_{\bp}, \bft^{v}_{\bp}$ and outputs the fine-level log differential opacity $\sigma_\bp$ and color $\bc_\bp$. 

For the image encoder and volume decoder, please refer to Table~\ref{tab:enc} and Table~\ref{tab:dec} for their architecture. 
To better model the view-dependent effects, we employ two decoders to regress the color and opacity at the coarse level. 
The common structure of each decoder is shown in Table~\ref{tab:dec}.
For the color decoder, the input is the concatenation of the latent vector $\bz{\in}\mathbb{R}^{256}$ and the camera view direction $\bv{\in}\mathbb{R}^{3}$, thus the final input size is $N^c_{in}=256+3$ 
and the output size is $N^c_{out}=3$, with a parallel branch producing view-dependent spatial scene features $\bft^{v}_{\bp}{\in}\mathbb{R}^{32}$.
Similarly, the opacity decoder only takes the latent vector $\bz{\in}\mathbb{R}^{256}$ as input and regresses opacity $\sigma_p{\in}\mathbb{R}$ and view-independent spatial scene features $\bft_{\bp}{\in}\mathbb{R}^{32}$ from its two branches respectively. 
%
To restrict the regressed color $\wt{\bc}_{\bp}$ to be non-negative, we apply a ReLU function after the last layer that directly outputs it. 

\begin{figure*}[!ht]
\begin{center}
\includegraphics[width=0.8\textwidth]{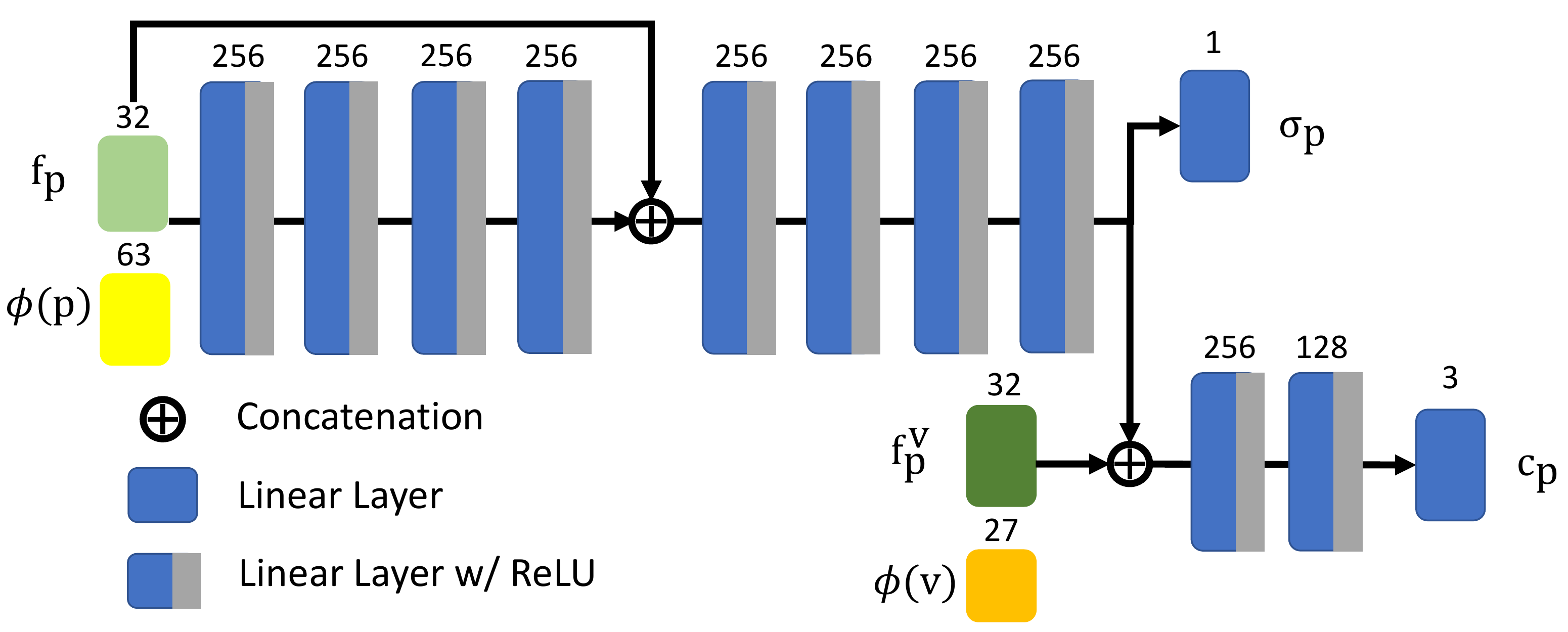}
\end{center}
\label{fig:mlp}
\caption{\textbf{Refinement MLP architecture}. 
Each blue box is a fully-connected layer and the number on top of each box is the output size of that layer. Blue box with gray tail is a linear layer with ReLU activation. Boxes in other color stand for different inputs and the size is marked on top. 
}
\end{figure*}

%
In Figure~\ref{fig:mlp}, we show the structure of the Refinement MLP. 
The spatial scene features $\bft_{\bp}, \bft^{v}_{\bp}$ are extracted from 
the feature voxel $\mathbf{V}_{\bp}$ with a continuous coordinate $\bp \in \bR^3$ using tri-linear interpolation. Log differential opacity $\sigma_\bp \in \bR$ is regressed from the last fully-connected layer of the top branch and no non-linearity is applied. 
The spatial color value $\bc_\bp$ is the output of the bottom branch and ReLU is applied afterwards to guarantee the regressed value is non-negative. At the beginning of the refinement network, a concatenation of a positional encoding of position $\bp$ and its corresponding view-independent spatial scene feature $\bft_{\bp}$. The color branch network learns to explain view-dependent effects by having additional inputs in addition to the positional encoding, such as the camera view $\bv$ and view-dependent spatial scene feature $\bft^{v}_{\bp}$ at position $\bp$. 
Note that the adapted version of NeRF, which we us as a baseline, shares
exactly the same architecture as shown in Figure~\ref{fig:mlp}, except that instead
of $\bft_{\bp}, \bft^{v}_{\bp}$ it uses the global latent vector $\bz$ as additional input.
 
\begin{table}[!ht]
\centering
\begin{tabular}{|c|c|c|}
\hline
   & \multicolumn{2}{c|}{Encoder}              \\ \hline
1  & \multicolumn{2}{c|}{\ttt{Conv2d}(9, 32)}        \\ \hline
2  & \multicolumn{2}{c|}{\ttt{Conv2d}(32, 64)}       \\ \hline
3  & \multicolumn{2}{c|}{\ttt{Conv2d}(64, 128)}      \\ \hline
4  & \multicolumn{2}{c|}{\ttt{Conv2d}(128, 128)}     \\ \hline
5  & \multicolumn{2}{c|}{\ttt{Conv2d}(128, 256)}     \\ \hline
6  & \multicolumn{2}{c|}{\ttt{Conv2d}(256, 256)}     \\ \hline
7  & \multicolumn{2}{c|}{\ttt{Conv2d}(256, 256)}     \\ \hline
8  & \multicolumn{2}{c|}{\ttt{Flatten}()}            \\ \hline
9  & \multicolumn{2}{c|}{\ttt{Linear}(256x4x2, 512)} \\ \hline
10 & \ttt{Linear}(512, 256)     & \ttt{Linear}(512, 256)   \\ \hline
\end{tabular}
\vspace{0.1cm}
\caption{\label{tab:enc}\textbf{Encoder architecture}. Each \ttt{Conv2d} layer in the encoder has a kernel size of $4$, stride of $2$ and padding of $1$. After each layer, except for the last two parallel fully-connected layers, a Leaky ReLU~\cite{maas2013leakyrelu} activation with a negative slope of $0.2$ is applied. The last two parallel fully-connected layers produce,
respectively, $\bm\mu$ and $\bm\sigma$.
}
\end{table}

\begin{table}[!ht]
\centering
\begin{tabular}{|c|c|c|}
\hline
  & \multicolumn{2}{c|}{Decoder}                            \\ \hline
1 & \multicolumn{2}{c|}{\ttt{Linear}($N^{X}_{in}$, 1024)}              \\ \hline
2 & \multicolumn{2}{c|}{\ttt{Reshape}(1024, 1, 1, 1)}             \\ \hline
3 & \ttt{ConvTrans3d}(1024, 512) & \ttt{ConvTrans3d}(1024, 512) \\ \hline
4 & \ttt{ConvTrans3d}(512, 512)  & \ttt{ConvTrans3d}(512, 512)  \\ \hline
5 & \ttt{ConvTrans3d}(512, 256)  & \ttt{ConvTrans3d}(512, 256)  \\ \hline
6 & \ttt{ConvTrans3d}(256, 256)  & \ttt{ConvTrans3d}(256, 256)  \\ \hline
7 & \ttt{ConvTrans3d}(256, 128)  & \ttt{ConvTrans3d}(256, 128)  \\ \hline
8 & \ttt{ConvTrans3d}(128, $N^{X}_{out}$)  & \ttt{ConvTrans3d}(128, 32)   \\ \hline
\end{tabular}
\vspace{0.1cm}
\caption{\label{tab:dec}\textbf{Decoder architecture}. Each layer is followed by a Leaky ReLU~\cite{maas2013leakyrelu} activation with a negative slope of $0.2$ except for the last two parallel layers. Each \texttt{ConvTrans3d} layer has a kernel size of $4$, a stride of $2$ and a padding of $1$. $N^{X}_{in}$ stands for the input feature size and
$N^{X}_{out}$ is the output size. $X$ here is a placeholder for color or opacity, $X\in \{c, \sigma\}$. 
%
}
\end{table}

\subsection{Hyperparameter Settings}

We use Adam~\cite{kingma2014adam} with a learning rate $1\mathrm{e-}4$, and $\beta_1=0.9, \beta_2=0.999$. All the models are trained for approximately $70-100\mathrm{K}$ iterations, 
each batch containing $64\times64$ rays. For each ray, we then uniformly sample $128$ 
query locations for the coarse level, and $32$ more locations for the fine level using our 
sampling scheme.
We set $\lambda_f=0.1, \lambda_c=0.1$ and $\lambda_{KL}=0.001$
Training on a sequence of 360 frames under $93$ camera views with $1024\times667$ resolution takes approximately 3-4 days on a single NVidia-V100-32GB GPU.
All our models are implemented in PyTorch.

\section{Novel View Synthesis}


\subsection{Qualitative Results}

\begin{figure*}[!ht]
    \centering
    \includegraphics[width=0.9\textwidth]{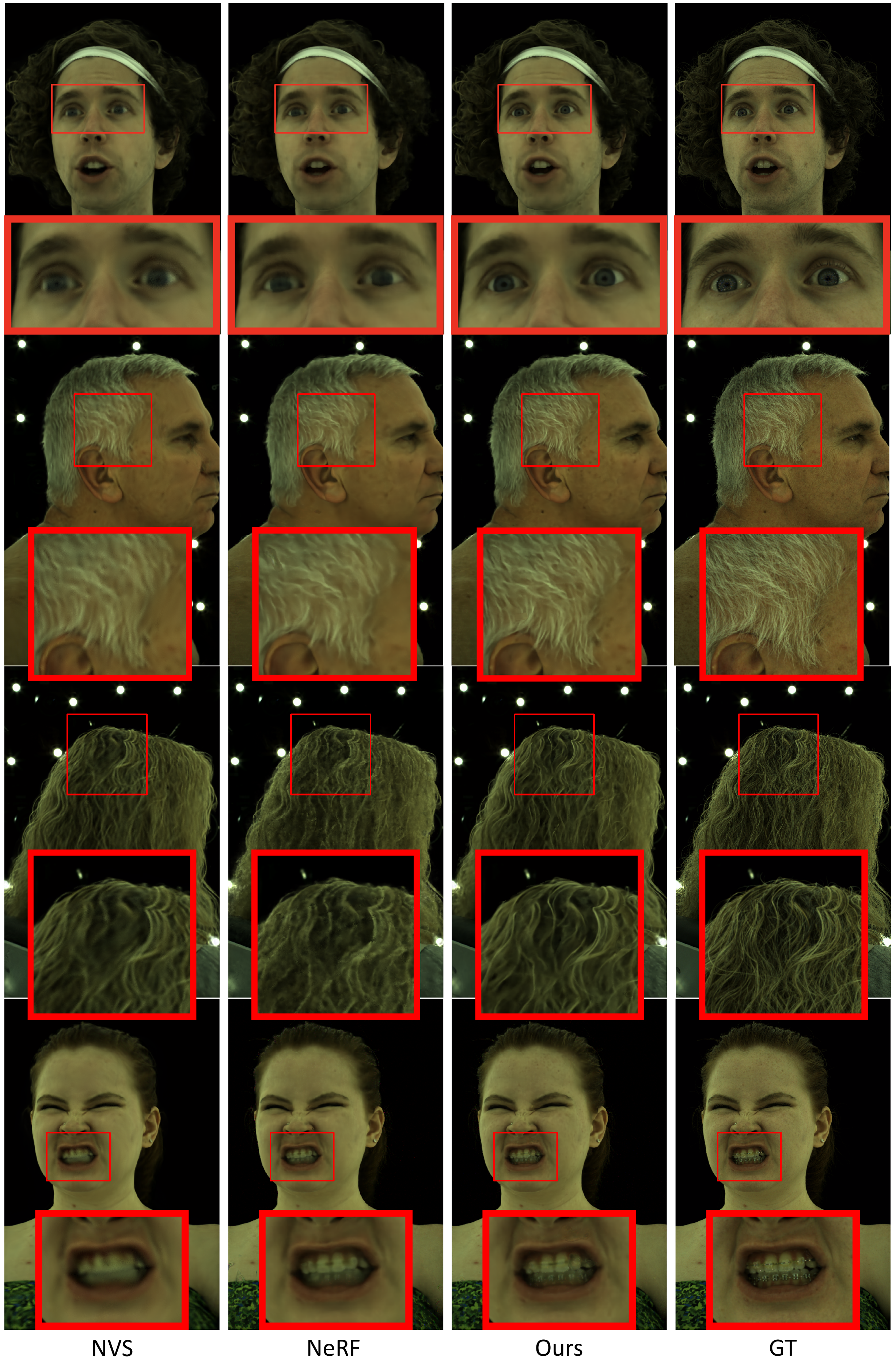}
    \caption{
    \textbf{Qualitative comparison of rendered images}.
    } 
    \label{fig:app_vis1}
\end{figure*}

We show more qualitative results in a larger size than in the main document in Figure~\ref{fig:app_vis1} and Figure~\ref{fig:app_vis2}. 
Please also refer to \texttt{videos/rot\_zoom} for more video results.

\begin{figure*}[!ht]
    \centering
    \vspace{-0.8cm}
    \includegraphics[width=0.9\textwidth]{pics/vis_fig2new.pdf}
    \caption{
    \textbf{Qualitative comparison of rendered images}.
    } 
    \label{fig:app_vis2}
\end{figure*}

\section{Animation}


\subsection{Latent Space Sampling}

We show more results of expression sampling in Figure~\ref{fig:app_latent_sample} and Figure~\ref{fig:app_latent_sample_aut}. Please see \texttt{videos/sample\_interp\_latent/interp\_X.mp4} and \texttt{videos/sample\_interp\_latent/sample\_X.mp4} for video results of sampling in latent space and interpolation. \texttt{X} here could be \textit{subj1} or \textit{subj2}. The first video contains 12 uniform keyframe expressions that are directly sampled from the latent space. Then, between each keyframe, we linearly interpolate 10 more frames to create the video. The second videos contains free view rendering of several sampled expressions.

\begin{figure*}[!ht]
    \centering
    \includegraphics[width=1.0\textwidth]{pics/sample_cat_app.png}
    \caption{
    \textbf{Rendering results of direct sampling in latent space}.
    } 
    \label{fig:app_latent_sample}
\end{figure*}

\begin{figure*}[!ht]
    \centering
    \includegraphics[width=0.8\textwidth]{pics/sample_cat_app_aut.png}
    \caption{
    \textbf{Rendering results of direct sampling in latent space}.
    } 
    \label{fig:app_latent_sample_aut}
\end{figure*}

\subsection{Landmark-driven Animation}
We used a PointNet~\cite{qi2017pointnet}-like encoder as 
a base architecture for the keypoint encoder.
Compared to the original work, our inputs are different in three aspects: 
1) The points are in 2D, 
2) The order of each point is fixed rather than arbitrary, 
3) All points are roughly aligned to a canonical pose.
To simplify the problem, we use the T-Net in the PointNet as the 
encoder that regresses the latent code from a set of points.
We show the architecture in Table~\ref{tab:kpsenc}.
More results of keypoint-driven animation can be found in Figure~\ref{fig:app_kps_render}. Please refer to 
\texttt{videos/kps\_render/subj1.mp4} more video results. In the video, the 2d keypoints in the blue bounding box are used as input to the keypoint encoder. The image in the middle is the output of our model and the image on the right most column is the ground truth. As we can see, the decoder in our method can also be driven by inputs from other modalities.

\begin{table}[!ht]
\centering
\begin{tabular}{|c|c|}
\hline
   & Kps Encoder       \\ \hline
1  & \ttt{Conv1d}(2, 64)     \\ \hline
2  & \ttt{Conv1d}(64, 128)   \\ \hline
3  & \ttt{Conv1d}(128, 256)  \\ \hline
4  & \ttt{Conv1d}(256, 512)  \\ \hline
5  & \ttt{Conv1d}(512, 1024) \\ \hline
6  & \ttt{MaxPool1d}()       \\ \hline
7  & \ttt{Flatten}()         \\ \hline
8  & \ttt{Linear}(1024, 512) \\ \hline
9  & \ttt{Linear}(512, 512)  \\ \hline
10 & \ttt{Linear}(512, 256)  \\ \hline
\end{tabular}
\vspace{0.1cm}
\caption{\label{tab:kpsenc}\textbf{Keypoint Encoder architecture}. Each layer is followed by a ReLU except for the last fully-connected layer. Each \ttt{Conv1d} layer has a kernel size of $1$, a stride of $1$ and a padding of $0$.}
\end{table}

\begin{figure*}[!ht]
    \centering
    \includegraphics[width=1.0\textwidth]{pics/kps_cat_app.png}
    \caption{
    \textbf{Keypoint-driven animation}.
    } 
    \label{fig:app_kps_render}
\end{figure*}

\subsection{Fitting New Sequences}

Please see \texttt{videos/sequence\_fitting/X\_noft.mp4}  for rendering results of the model without finetuning and \texttt{videos/sequence\_fitting/X\_enc.mp4} for rendering results of the model with encoder-only finetuning. \texttt{X} could be either 
\textit{subj1} or \textit{subj2}. In both videos, the images in the red bounding boxes serve as inputs. The image in the middle is the output of our model and the image on the right most column is the ground truth. As we can see, the model without finetuning can achieve reasonable performance on fitting the new sequence. And with only encoder finetuning, the encoder quickly adapts to the latent space of the decoder on the novel sequence and creates much smoother results. 
For free view rendering results of both models, please see \texttt{videos/sequence\_fitting/X\_enc\_fr.mp4} and \texttt{videos/sequence\_fitting/X\_noft\_fr.mp4}

{\small
\bibliographystyle{ieee_fullname}
\bibliography{egbib}
}